\documentclass[10pt,twocolumn,letterpaper]{article}

\usepackage{cvpr}
\usepackage{times}
\usepackage{epsfig}
\usepackage{graphicx}
\usepackage{amsmath}
\usepackage{amssymb}
\usepackage{subfig}
\usepackage{algpseudocode}      
\usepackage{algorithm}          
\usepackage{gensymb}            

\usepackage[pagebackref=true,breaklinks=true,letterpaper=true,colorlinks,bookmarks=false]{hyperref}

\cvprfinalcopy 

\def\cvprPaperID{****} 

\ifcvprfinal\fi
\begin{document}

\title{Guided Labeling using Convolutional Neural Networks}

\author{Sebastian Stabinger\\
  {\tt\small Sebastian.Stabinger@uibk.ac.at}
  \and
  Antonio Rodr\'iguez-S\'anchez\\
  {\tt\small Antonio.Rodriguez-Sanchez@uibk.ac.at}
  \and
  University of Innsbruck\\
  Technikerstrasse 21a, Innsbruck, Austria\\
}

\maketitle

\begin{abstract}
  Over the last couple of years, deep learning and especially
  convolutional neural networks have become one of the work horses of
  computer vision. One limiting factor for the applicability of
  supervised deep learning to more areas is the need for large,
  manually labeled datasets. In this paper we propose an easy to
  implement method we call guided labeling, which automatically
  determines which samples from an unlabeled dataset should be
  labeled. We show that using this procedure, the amount of samples
  that need to be labeled is reduced considerably in comparison to
  labeling images arbitrarily.
\end{abstract}

\section{Introduction}
Deep learning has gained a lot of interest over the last few years
because the methods perform very well on a wide range of machine
learning tasks. One class of especially successful deep learning
methods are convolutional neural networks (CNNs) for image
classification.

Unfortunately, CNNs need a large amount of labeled training data to
perform well. In many cases, this labeling is performed by humans. A
common approach is to use some form of crowd based labeling. For
example, Amazon Mechanical Turk~\cite{turk2012amazon} was used for
labeling the ImageNet dataset~\cite{deng2009imagenet}. Data can also
be obtained as a side effect of some human interaction with an online
system. For example, CAPTCHA~\cite{von2003captcha} challenges to
prevent bots using online services can be set up to produce labeled
data as a side effect of the verification
procedure~\cite{faymonville2009captcha}.

Alas, simply labeling all available samples is a very inefficient use
of human labor, since not all samples will be of equal value. On the
one hand, adding a sample which is similar to samples already in the
dataset will not be very usefull. On the other hand, in most cases,
not all classes will have the same difficulty and it might make sense
to add more samples of the difficult classes to the dataset.
Therefore, it would be advantageous to label a sample which would
maximize the classification accuracy of a system. Unfortunately, how
much the quality of a dataset would increase by adding a specific
labeled sample can only be determined after labeling and training on
the resulting dataset. This is obviously not useful if we want to
decide which samples should be labeled in the first place.

We propose to use the classification confidence of a CNN while trying
to predict the class of unlabeled images to decide what to label next.
The proposed procedure, together with extensive data augmentation,
will be evaluated for two small neural networks on the
MNIST~\cite{lecun1998gradient} and
CIFAR10~\cite{krizhevsky2009learning} dataset.

In practice we propose the following workflow: A small, labeled
dataset is used to train a neural network that is used to select a
batch of the most confusing images from a set of unlabeled data. This
batch is given to human workers who label the images, after which they
are added to the training dataset and the process repeats. We call
this procedure \emph{guided labeling}. Our hypothesis is that, using
this procedure, we are able to trade human for computational
resources.

\begin{figure}[t]
  \centering
  \includegraphics[width=0.45\textwidth]{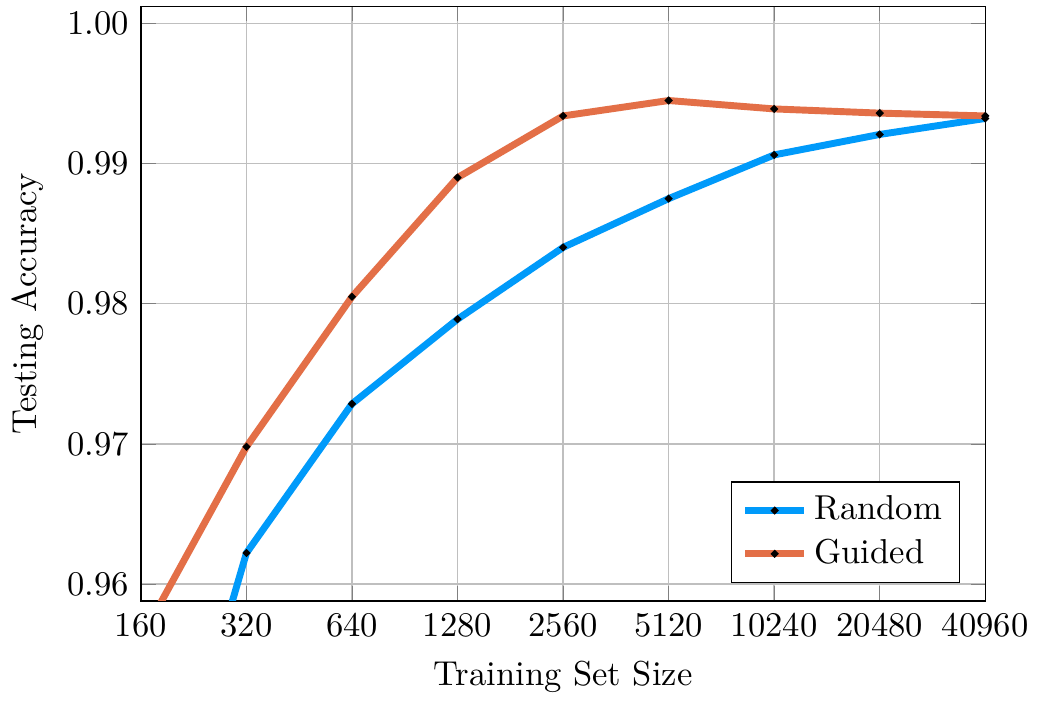}
  \caption{Classification accuracy for the MNIST dataset depending on
    the size of the training set. It was either randomly sampled from
    the dataset or generated using the proposed guided labeling
    approach. Note that the training set size is on a
    \textit{logarithmic scale}.\label{mnistacc}}
\end{figure}

\section{Related Work}
\label{sec:related-work-1}

The idea that a system can decide for itself which data it wants to
learn from is generally called \textit{active learning}. A survey of
the field has been published by Settles\cite{settles2010active}. The
concept of active learning has already been used in 1988 by
Angluin~\cite{angluin1988queries}, although in this case the samples
to be labeled were not chosen from a preexisting unlabeled dataset,
but synthetically generated by the learner itself. This method is
currently not feasible for image classification. Such a method has
been shown in 1992 by Baum and Lang~\cite{baum1992query} to work
poorly for handwritten character recognition.

Active learning, where a small amount of labeled data and a larger,
fixed set of unlabeled data is available is generally called
\textit{pool--based sampling} as first presented in 1994 by Lewis
\etal~\cite{lewis1994sequential} for learning text classifiers.
Pool--based sampling has been used for the task of image classification
in 2001 by Tong \etal~\cite{tong2001support} using support vector
machines.

The idea of using the uncertainty of a system to guide the labeling is
called \textit{uncertainty sampling} in the field of active learning
and was also already introduced in 1994 by Lewis
\etal~\cite{lewis1994sequential} although they did not use the entropy
of a resulting probability distribution as we propose for our method.
Entropy as a confusion measure has been used in 2004 by Rebecca
Hwa~\cite{hwa2004sample} for statistical parsing.

\section{Methodology}
\label{sec:empl-arch}
The following network architectures were used for the experiments.
Note that the networks were not selected to give the best possible
results for the specific dataset. The important aspect is the
difference between the performance on randomly selected images and
images selected by guided labeling. All layers, except for the last
one, use a ReLU~\cite{nair2010rectified} activation function. The last
layer uses a softmax activation function.

For the MNIST dataset we employed a network consisting of the
following seven layers. Starting from the input layer we have a
convolutional layer with 64 $3 \times 3$ kernels, a convolutional layer
with 128 $3 \times 3$ kernels, a max pooling layer with a pooling size of
$2 \times 2$, a dropout layer~\cite{srivastava2014dropout} with a dropout
probability of 0.25, a fully connected layer with 128 output neurons,
a dropout layer with a dropout probability of 0.5 and a fully
connected layer with 10 output neurons as the last and output layer.

For the CIFAR10 dataset we employ a network consisting of the
following eleven layers. A convolutional layer with 32 $3 \times 3$ kernels
as the input layer, a convolutional layer with 32 $3 \times 3$ kernels, a
maxpooling layer with a pooling size of $2 \times 2$, a dropout layer with
a dropout probability of 0.25, a convolutional layer with 64
$3 \times 3$ kernels, a convolutional layer with 64 $3 \times 3$ kernels, a
maxpooling layer with a pooling size $2 \times 2$, a dropout layer with a
dropout probability of 0.25, a fully connected layer with 512 output
neurons, a dropout layer with a dropout probability of 0.5, and a
fully connected layer with 10 output neurons as the output layer.
\subsection{Data Augmentation}
\label{sec-2-2}
During training, the dataset is randomly augmented for each training
epoch. Thus, the network never sees two identical images during
training.

For the MNIST dataset, the following augmentations are performed: A
rotation in the range $\pm 25 \degree$ is applied, the image is scaled
in the range $0.8$ to $1.2$, the image is sheared on the X and Y axis
in a range of $\pm 20$ pixels, a random elastic distortion as
presented by Simard et al. \cite{simard2003best} is applied, and the
image is cropped back to the original size of $28 \times 28$ pixels if
necessary.

For the CIFAR10 dataset, the images are randomly mirrored along the
vertical axis, are rotated in the range $\pm 25 \degree$, are scaled
in the range $0.8$ to $1.2$, and are cropped to the original size of
$32 \times 32$ pixels.
\subsection{Measuring Confusion of the Network}
\label{sec:conf}
Following the work of Park et al. \cite{park2015using} for other
machine learning methods, we use the \emph{response distribution
  entropy (RDE)} as a classification confidence measure. Feeding a
sample through a classification network gives us a probability
distribution over all possible classes, which will be called the
response distribution. The entropy~\cite{shannon2001mathematical} of
this distribution serves as a measure of the overall certainty of the
network regarding the classification.

Given a categorical probability distribution with a set of possible
classes $C$ and a probability of class $c \in C$ given by $P(c)$, the
entropy of this distribution can be calculated by:
\begin{equation}
  \label{eq:1}
  H(C) = -\sum_{c \in C} P(c) \log_2 P(c)
\end{equation}
If a system predicts a single class with a probability of $1$ this
gives us an RDE of $0$ bits. A completely uniform probability
distribution over N classes gives us an RDE of $log_2 |C|$ bits with
$|C|$ being the number of classes. Thus, the confidence of a network
in the classification decreases with an increasing response
distribution entropy.

\subsection{Dataset Imbalance}
\label{sec-2-5}
One problem that might arise from generating a dataset by selecting
the most confusing samples is that it might become very unbalanced
since the most difficult classes will become overrepresented. As shown
in the experimental section, this is what happens in practice. We
assume that this is, up to a certain point, a beneficial side effect
of the guided labeling procedure since we want more samples of
confusing classes in our dataset. Unfortunately, extremely unbalanced
datasets are problematic when training a neural network. To lessen the
effects of imbalance, we weigh misclassifications of underrepresented
classes higher in the loss function during training. The weights are
calculated in the following fashion:

\begin{equation}
  \label{eq:2}
  \text{weight}(c) = \max\left(\log\left(\frac{\mu t}{c}\right), 1.0\right)
\end{equation}

where $t$ is the total number of samples in the training dataset, $c$
is the number of samples for a specific class in the dataset and $\mu$
is a scaling factor which has to be determined empirically. We set
$\mu = 0.3$ as it provided the best results in our experimental
evaluation.

\subsection{The Guided Labeling Algorithm}
\label{sec-2-4}
Training starts with a small, labeled training set and a large set of
unlabeled images. Depending on the used dataset, one of the
convolutional neural networks from section \ref{sec:empl-arch} is
trained on the training set.

The images of the unlabeled dataset are augmented by the procedure
presented in section \ref{sec-2-2}. The augmented images are fed into
the trained network, which returns class probabilities (\ie the
response distribution). The entropy for this distribution is
calculated according to section \ref{sec:conf}, giving us the response
distribution entropy for one augmentation of one image. This procedure
is repeated multiple times for different augmentations and the average
RDE for each image is recorded. This average RDE is an indication for
how confusing a certain unlabeled image is for the trained network,
also accounting for the employed data augmentation.

A predetermined amount of the most confusing images (the images with
the highest average RDE) are selected for labeling. Those images are
removed from the unlabeled dataset, labeled by a human, and added to
the training set. The CNN is then retrained on this new dataset.

This procedure is repeated until a satisfactory performance is
reached, or there is no data left to be labeled. Our hypothesis is
that, using this procedure instead of randomly labeling images,
satisfactory performance is reached with a dataset containing fewer
labeled samples. Algorithm \ref{alg1} presents pseudo code for the
guided labeling procedure.

\begin{algorithm*}
  \begin{algorithmic}
    \Procedure{LabelConfusingSamples}{$m,X_u,n$} \Comment{$m$ CNN model, $X_u$ unlabeled samples, $n$ \# of images to label}
    \ForAll{$x \in X_u$}
      \State $R \leftarrow \emptyset$ \Comment{Set to calculate mean RDE for each image}
      \For{$j \leftarrow 1 \text{ to } k$} \Comment{$k$ determines how many augmentations are used for calculating the RDE}
        \State $x_a \leftarrow \text{augment}(x)$
        \State $R \leftarrow R \cup \text{RDE}(m,x_a)$ \Comment{Get response distribution entropy for augmented sample $x_u$ and add to set}
      \EndFor
      \State $r_x \leftarrow \text{mean}(R)$  \Comment{Calculate mean RDE for image $x$}
    \EndFor
    \State $X_l \leftarrow $ select $n$ samples with highest $r_x$
    \State $Y_l \leftarrow $ get labels for $X_l$ by manual labeling
    \State \Return $X_l, Y_l$
    \EndProcedure
    \Statex
    \Procedure{GuidedLabeling}{$X,Y,X_u$}\Comment{$X$ samples and $Y$ related labels, $X_u$ unlabeled samples }
    \While{achieved accuracy not sufficient $\wedge$ $|X_u| \ge |X|$}
      \State $m \leftarrow \text{random}()$ \Comment{Randomly Initialize the CNN model}
      \While{model performance still improves on validation dataset}
        \State $X_a \leftarrow \text{augment}(X)$ \Comment{Augment images with random perturbations}
        \State $m \leftarrow \text{trainCNN}(m,X_a,Y)$ \Comment{Train CNN for one epoch}
        \EndWhile{}
        \State $X_l,Y_l \leftarrow $\Call{LabelConfusingSamples}{$m,X_u,|X|$} \Comment{$|X|$ = number of samples in $X$}
        \State $X \leftarrow X \cup X_l, Y \leftarrow Y \cup Y_l$ \Comment{Add selected samples to labeled dataset}
        \State $X_u \leftarrow X_u \backslash X_l$ \Comment{Remove selected samples from unlabeled dataset}
      \EndWhile{}
    \State \Return{$X,Y$} \Comment{Return the new labeled dataset}
    \EndProcedure
  \end{algorithmic}
  \caption{Guided Labeling Procedure\label{alg1}}
\end{algorithm*}

\section{Experimental Evaluation}
\label{sec-3}
There are two problems with evaluating the proposed guided labeling
method when actually using humans for labeling. First, the procedure
is very time consuming and needs a lot of human resources if different
modalities have to be tested. Second, it is very hard to decide
whether the guided labeling performs better than just randomly
labeling images. Because of this, we decided to run the experiments in
the following fashion: The network is trained on a small subset of the
datasets (MNIST and CIFAR10 respectively) and the remaining images are
treaded as being unlabeled. The guided labeling procedure selects
images to be labeled, after which they are added to the training set
using the labels provided by the dataset.

\subsection{Training of the Networks}
\label{sec-3-1}
All our experiments used ADAM \cite{kingma2014adam} as the optimizer
with a learning rate of $0.01$ and categorical cross--entropy as the
loss function. All weights were initialized using normalized
initialization by Glorot \etal~\cite{glorot2010understanding}.

During training, we utilized early stopping, using the accuracy of a
validation dataset of $10,000$ images as the stopping criterion. A
patience of $100$ epochs was used. This means that if there was no
improvement of the accuracy on the validation dataset for $100$
epochs, training was stopped, and the network from $100$ epochs ago is
used for testing purposes.
\subsection{Selecting the Amount of Images to be Labeled}
\label{sec-3-2}
When using the guided labeling approach, one has to decide how many
images should be selected for labeling in each step. In principle, one
would expect the method to work better if fewer images are selected in
each iteration, since the newly labeled images can be used to inform
the selection of images in the next iteration. On the other hand,
selecting fewer images results in more iterations. Since for each
iteration a neural network has to be trained, the procedure takes
longer. In addition, fewer images also means that the labeling
workflow including human labor is less efficient. We determined, that
an exponential selection scheme offers a good compromise. In the
exponential selection scheme, as many images are selected for labeling
as are already in the training set. Thus, in every iteration the size
of the training set doubles. The number of images to label in
iteration $i$ when starting with a labeled training set of size $s$
can be calculated as follows:

\begin{equation}
  \label{eq:3}
  2^{i-1}s
\end{equation}

\section{Results}
\label{sec-4}
For comparison reasons, identical network architectures are also
trained on an equally sized and randomly selected subset of the given
dataset. The prediction accuracy of the network trained on images
selected using guided labeling and the network trained on randomly
selected images is compared on the testing sets provided with the
MNIST and CIFAR10 dataset respectively. Prediction accuracy is the
number of correctly classified images in relation to the number of
tested images.
\subsection{MNIST}
\label{sec-4-1}
A comparison of guided labeling using the exponential selection scheme
and randomly selected images can be seen in \autoref{mnistacc} for the
MNIST dataset. Note, that the training set size is on a logarithmic
scale. It is obvious, that guided labeling is very helpful for MNIST.
Not only are we able to achieve the same accuracy with a $16$ times
smaller training set ($2,560$ instead of $40,960$ images), but $5,120$
selected images even outperfroms the use of the whole dataset.

An important question is, whether the response distribution entropy of
images does actually correlate with samples that would be considered
more or less difficult by a human observer. \autoref{mnistselected}
shows the twenty most confusing (highest RDE) and least confusing
(lowest RDE) images during the second guided labeling iteration. The
most confusing images are clearly difficult images. Many are either
unconventional writing styles of numbers, are of poor quality or
contain noise. On the other hand, the twenty images with the lowest
RDE all show slightly different variations of the number 3 which would
probably not add much to the training set, considering that data
augmentation is used during training.

\def\imgsize{0.041\textwidth}
\begin{figure}[hptb]
  \centering
  \subfloat{
    \includegraphics[width=\imgsize]{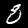}
    \includegraphics[width=\imgsize]{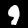}
    \includegraphics[width=\imgsize]{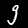}
    \includegraphics[width=\imgsize]{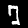}
    \includegraphics[width=\imgsize]{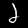}
    \includegraphics[width=\imgsize]{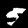}
    \includegraphics[width=\imgsize]{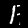}
    \includegraphics[width=\imgsize]{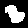}
    \includegraphics[width=\imgsize]{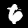}
    \includegraphics[width=\imgsize]{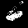}
  }

  \subfloat[a][Twenty most confusing images]{
    \includegraphics[width=\imgsize]{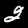}
    \includegraphics[width=\imgsize]{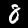}
    \includegraphics[width=\imgsize]{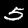}
    \includegraphics[width=\imgsize]{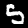}
    \includegraphics[width=\imgsize]{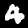}
    \includegraphics[width=\imgsize]{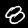}
    \includegraphics[width=\imgsize]{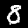}
    \includegraphics[width=\imgsize]{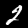}
    \includegraphics[width=\imgsize]{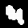}
    \includegraphics[width=\imgsize]{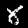}
  }

  \subfloat{
    \includegraphics[width=\imgsize]{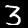}
    \includegraphics[width=\imgsize]{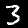}
    \includegraphics[width=\imgsize]{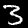}
    \includegraphics[width=\imgsize]{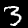}
    \includegraphics[width=\imgsize]{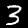}
    \includegraphics[width=\imgsize]{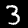}
    \includegraphics[width=\imgsize]{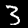}
    \includegraphics[width=\imgsize]{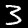}
    \includegraphics[width=\imgsize]{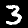}
    \includegraphics[width=\imgsize]{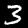}
  }
  
  \subfloat[a][Twenty least confusing images]{
    \includegraphics[width=\imgsize]{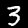}
    \includegraphics[width=\imgsize]{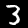}
    \includegraphics[width=\imgsize]{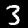}
    \includegraphics[width=\imgsize]{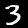}
    \includegraphics[width=\imgsize]{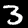}
    \includegraphics[width=\imgsize]{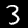}
    \includegraphics[width=\imgsize]{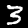}
    \includegraphics[width=\imgsize]{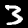}
    \includegraphics[width=\imgsize]{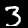}
    \includegraphics[width=\imgsize]{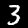}
  }

  \caption{The twenty most and least confusing images of the MNIST
    dataset, selected by the response distribution entropy after two
    guided labeling steps started on 160 randomly selected
    images.\label{mnistselected}}
\end{figure}

\begin{figure}
  \centering \subfloat[Distribution of the images in the training set
  among the possible classes, for different training set sizes. The
  training set size is increased using guided labeling. Each doubling
  in training set size corresponds to one iteration of the guided
  labeling procedure.]{ \def\svgwidth{0.4\textwidth} \footnotesize
    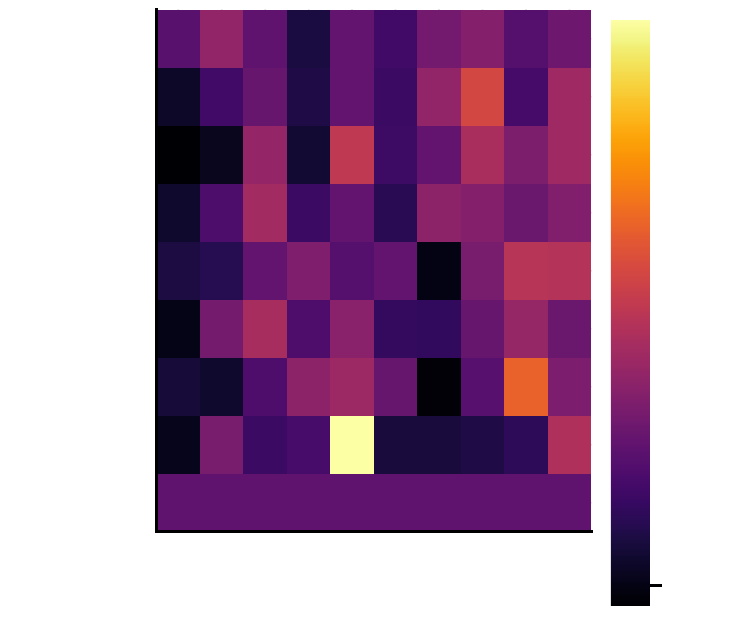 }
  
  \subfloat[Average distribution of the images in the training set
  among the possible classes over all training set sizes.]{
    \includegraphics[width=0.4\textwidth]{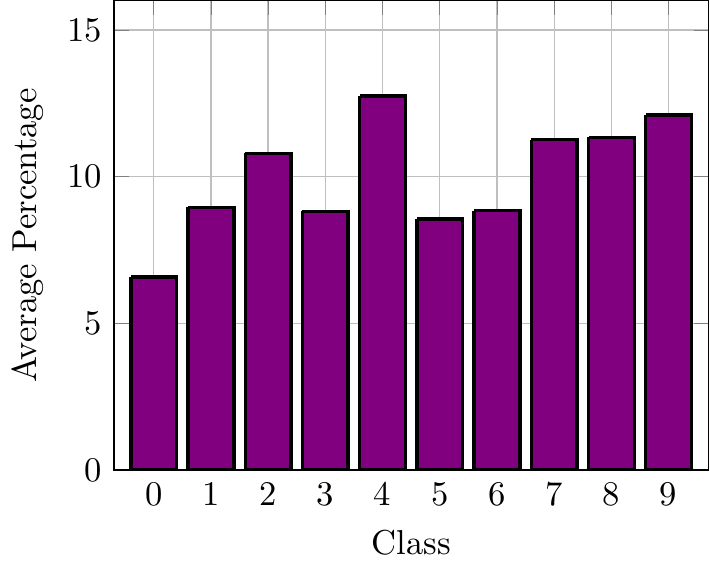} }
  \caption{Visualization of the class distribution for different
    training set sizes for the MNIST dataset. Guided labeling was used
    to increase the training set size.\label{mnistdist}}
\end{figure}

One could assume that the number 3 is just in general very distinctive
and not confusing for the CNN. To check this we can look at the class
distributions in the training set for each guided labeling iteration.
This is visualized in \autoref{mnistdist} in addition to the average
class distribution over all iterations, and it shows that the number
$3$ seems to be one of the less confusing numbers for the system.
Still, overall the MNIST dataset seems to be relatively balanced
regarding how confusing different classes are.

\subsection{CIFAR10}
\label{sec-4-2}
\begin{figure}
  \centering
  \includegraphics[width=0.45\textwidth]{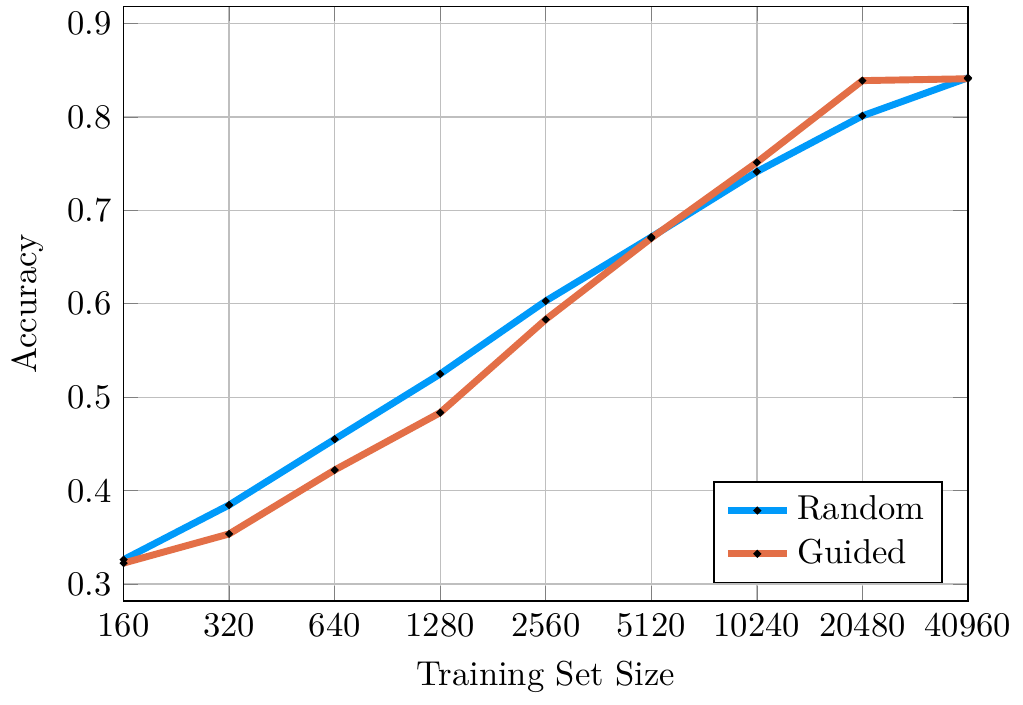}
  \caption{Classification accuracy for the CIFAR10 dataset depending on
    the size of the training set. It was either randomly sampled from
    the dataset or generated using the proposed guided labeling
    approach. Note that the training set size is on a
    \textit{logarithmic scale}.\label{cifar10acc}}
\end{figure}

\begin{figure}[pt]
  \centering
  \subfloat{
    \includegraphics[width=\imgsize]{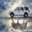}
    \includegraphics[width=\imgsize]{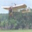}
    \includegraphics[width=\imgsize]{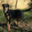}
    \includegraphics[width=\imgsize]{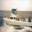}
    \includegraphics[width=\imgsize]{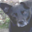}
    \includegraphics[width=\imgsize]{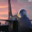}
    \includegraphics[width=\imgsize]{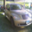}
    \includegraphics[width=\imgsize]{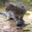}
    \includegraphics[width=\imgsize]{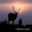}
    \includegraphics[width=\imgsize]{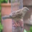}
  }

  \subfloat[Twenty most confusing images]{
    \includegraphics[width=\imgsize]{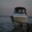}
    \includegraphics[width=\imgsize]{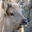}
    \includegraphics[width=\imgsize]{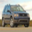}
    \includegraphics[width=\imgsize]{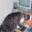}
    \includegraphics[width=\imgsize]{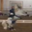}
    \includegraphics[width=\imgsize]{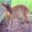}
    \includegraphics[width=\imgsize]{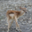}
    \includegraphics[width=\imgsize]{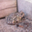}
    \includegraphics[width=\imgsize]{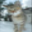}
    \includegraphics[width=\imgsize]{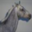}

  }

  \subfloat{
    \includegraphics[width=\imgsize]{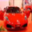}
    \includegraphics[width=\imgsize]{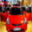}
    \includegraphics[width=\imgsize]{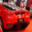}
    \includegraphics[width=\imgsize]{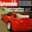}
    \includegraphics[width=\imgsize]{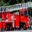}
    \includegraphics[width=\imgsize]{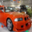}
    \includegraphics[width=\imgsize]{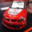}
    \includegraphics[width=\imgsize]{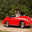}
    \includegraphics[width=\imgsize]{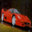}
    \includegraphics[width=\imgsize]{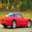}
  }
  
  \subfloat[Twenty least confusing images]{
    \includegraphics[width=\imgsize]{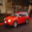}
    \includegraphics[width=\imgsize]{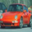}
    \includegraphics[width=\imgsize]{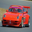}
    \includegraphics[width=\imgsize]{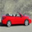}
    \includegraphics[width=\imgsize]{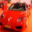}
    \includegraphics[width=\imgsize]{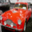}
    \includegraphics[width=\imgsize]{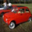}
    \includegraphics[width=\imgsize]{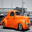}
    \includegraphics[width=\imgsize]{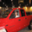}
    \includegraphics[width=\imgsize]{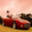}
  }

  \caption{The twenty most and least confusing images of the CIFAR10
    dataset, selected by the response distribution entropy after two
    guided labeling steps started on 160 randomly selected
    images.\label{cifar10selected}}
\end{figure}

CIFAR10 seems to be an overall more difficult dataset and the gains
from guided labeling are smaller, as can be seen in
\autoref{cifar10acc}. We give a possible reasons why this may be the
case in the discussion. Up until $5,120$ images, guided labeling even
performs worse than purely random selection. This presumably happens
because up to this point the training set consists of a lot of
confusing special cases which do not generalize well to the testing
set. After this point, guided labeling performs strictly better than
random selection until the training set reaches $40,960$ images, by
which point more or less the whole CIFAR10 dataset is used for
training. With guided labeling, almost the same performance is reached
with $20,480$ images as with using the whole dataset, cutting the
amount of images that have to be labeled in half.
Analogously to the MNIST dataset we also look at the twenty most and
least confusing images in \autoref{cifar10selected}. The twenty most
confusing images are a bit harder to interpred in this case, but
compared to the twenty least confusing images a clear difference can
be seen. The least confusing images almost exclusively contain bright
red cars. Presumably a color that does not appear in images of other
classes very often. Compared to the instances of cars in the most
confusing images it is obvious that the response distribution entropy
does correlate with the difficulty of the images.

\begin{figure}
  \centering
  \subfloat[Distribution of the images in the training set
  among the possible classes, for different training set sizes. The
  training set size is increased using guided labeling. Each doubling
  in training set size corresponds to one iteration of the guided
  labeling procedure.]{
    \def\svgwidth{0.4\textwidth}
    \footnotesize
    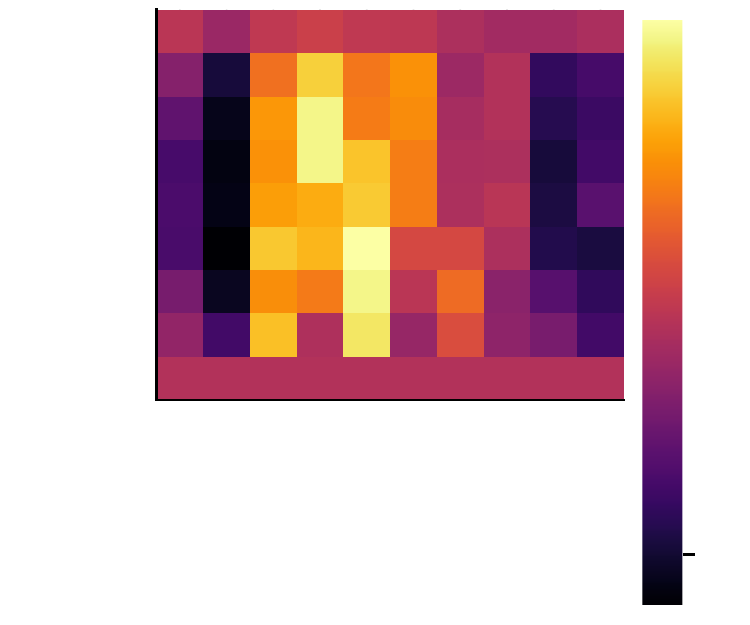
  }
  
  \subfloat[Average distribution of the images in the training set
  among the possible classes over all training set sizes.]{
    \includegraphics[width=0.4\textwidth]{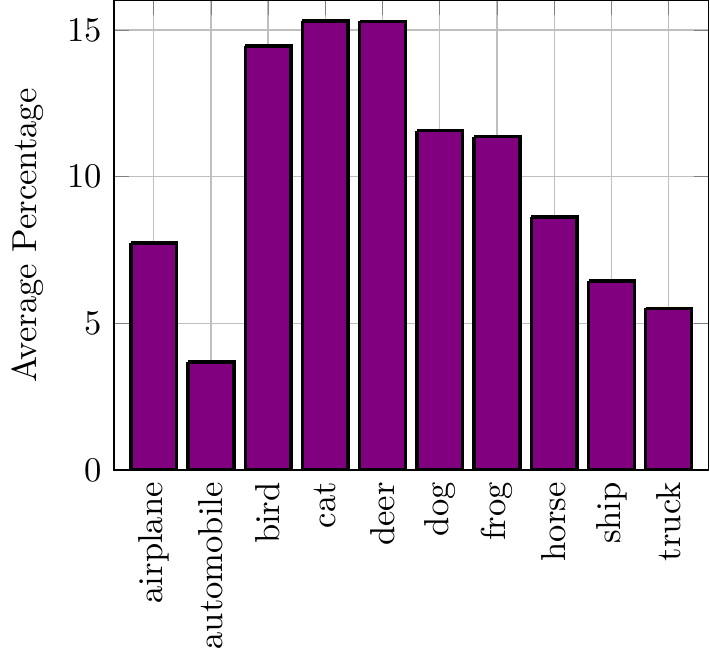} }
  \caption{Visualization of the class distribution for different
    training set sizes for the CIFAR10 dataset. Guided labeling was
    used to increase the training set size.\label{cifar10classdist}}
\end{figure}

Looking at the class distributions in \autoref{cifar10classdist} it is
clear that cars are the least confusing class. Presumably because the
overall color of the image is already very indicative of this class.
It is interesting to note that the technical classes (airplane,
automobile, ship, truck) are in general less confusing than the animal
classes (especially bird, cat, deer). Looking at the dataset, a lot of
the images of technical classes have a very distinct overall
appearence (e.g. airplanes usually have sky in the background, ships
have water, \dots). Thus, the training set can become very imbalanced
for CIFAR10. The class weights presented in Section \ref{sec-2-5} can
mitigate this imbalance to a certain degree.

\section{Discussion}
\label{sec-5}
First, we could show that the response distribution entropy is able to
identify difficult examples from an unlabeled dataset in case of the
MNIST as well as the CIFAR10 dataset. Even when trained on very few
images, the response distribution entropy of unlabeled data seems to
be a very good metric to detect redundant samples which do not need to
be labeled. Second, we demonstrated that the presented guided labeling
scheme itself can potentially reduce the number of samples that have
to be labeled by a large amount. In case of MNIST a well selected
dataset $\frac{1}{16}$ in size achieves the same classification
accuracy. For the CIFAR10 dataset the size could still be reduced by
one half. Still, why does the procedure perform worse on CIFAR10 than
on MNIST? We hypothesize two main reasons. On the one hand, the
dataset is much more difficult in general since the classes have much
higher variation, which can also not be replicated as easily by
augmentation of the images. For example, an image of a bird will
always have more variation than an image of the number 7. There is
more variation in the background, the lighting, different bird
species, etc. On the other hand, the CIFAR10 dataset is also much
smaller with respect to the difficulty and therefore less exhaustive.
It contains about the same amount of training data as MNIST, despite
being much more difficult.

This can also be seen when looking at the achieved accuracy depending
on the size of a randomly selected training set. Comparing
\autoref{mnistacc} for MNIST and \autoref{cifar10acc} for CIFAR10, it
is apparent that a bigger dataset would likely improve the accuracy
for CIFAR10, but would not for MNIST. Going from $20,480$ images to
$40,960$ images improved the accuracy by $\approx4\%$. This is about the
same improvement as going from $10,240$ to $20,480$ images. This
suggests that increasing the dataset size would likely further improve
the achievable accuracy. In case of MNIST, going from $20,480$ images
to $40,960$ images only improved the accuracy by $\approx0.1\%$. So
presumably the MNIST dataset is, using augmentation, more or less
exhaustive and additional training data would not improve the
achievable accuracy significantly.

Presumably, the presented active learning scheme works best if the set
of unlabeled data is very big and the achievable accuracy would
increase with the size of the unlabeled dataset without increasing the
number of samples that actually have to be labeled.

Future research should evaluate the presented method on additional
datasets(\eg CIFAR100~\cite{krizhevsky2009learning},
PASCAL~\cite{Everingham10} and ImageNet~\cite{ILSVRC15}). It will also
be interesting to see how well a dataset selected using one network
architecture will generalize to another architecture. We hypothesized
earlier that guided labeling likely would perform better for tasks
where a large number of unlabeled data is available. Experiments with
synthetically generated dataset, where the amount of unlabeled data is
unlimited and labels can easily be generated for each sample, might be
able to clarify this. It might also be interesting to bring additional
concepts from the field of active learning \cite{settles2010active} to
the area of deep learning.
\section{Conclusion}
We presented an active learning method we named \textit{guided
  labeling}, that allows for automatic selection of
confusing/difficult samples from a pool of unlabeled samples. The
method is easy to implement, yet we could show that using it reduces
the number of samples needed to achieve a desired accuracy
considerably. For MNIST the size of the dataset could be reduced 16
fold, for CIFAR10 it was cut in half. This might move manual labeling
to the region of feasibility for some tasks, especially since we
hypothesize that results would improve even more if the pool of
unlabeled data is very big. In case of the MNIST, the reduced dataset
even outperformed training on the full dataset, presumably by removing
unnecessary variations that hinder generalizability.

On a more general note, we could show that the field of active
learning might have a lot to offer to deep learning practitioners.

{\small
  \bibliographystyle{ieee}
  \bibliography{references}
}

\end{document}